\pdfoutput=1

\documentclass[11pt]{article}

\usepackage[preprint]{acl}

\usepackage{times}
\usepackage{latexsym}

\usepackage[T1]{fontenc}

\usepackage[utf8]{inputenc}

\usepackage{microtype}

\usepackage{inconsolata}

\usepackage{graphicx}
\usepackage{float}
\usepackage{amsmath,amsthm, mathtools, amssymb}
\newcommand{\bs}[1]{\mathbf{#1}}
\renewcommand{\figurename}{Figure}

\title{Statistical Uncertainty in Word Embeddings: GloVe-V}

\author{Andrea Vallebueno\thanks{Equal contribution.} \\
  Stanford University  \\
  \url{avaimar@stanford.edu} \\\And
  Cassandra Handan-Nader\footnotemark[1] \\
  Stanford University  \\
  \url{slnader@stanford.edu} \\\AND
  Christopher D. Manning \\
  Stanford University \\
  \url{manning@cs.stanford.edu}
  \\\And 
  Daniel E. Ho\thanks{Corresponding author.}   \\
  Stanford University \\
  \url{deho@stanford.edu}
  }

\begin{document}
\maketitle

\begin{abstract}
Static word embeddings are ubiquitous in computational social science applications and contribute to practical decision-making in a variety of fields including law and healthcare. However, assessing the statistical uncertainty in downstream conclusions drawn from word embedding statistics has remained challenging. When using only point estimates for embeddings, researchers have no streamlined way of assessing the degree to which their model selection criteria or scientific conclusions are subject to noise due to sparsity in the underlying data used to generate the embeddings. We introduce a method to obtain approximate, easy-to-use, and scalable reconstruction error variance estimates for GloVe \citep{pennington2014}, one of the most widely used word embedding models, using an analytical approximation to a multivariate normal model. To demonstrate the value of embeddings with variance (GloVe-V), we illustrate how our approach enables principled hypothesis testing in core word embedding tasks, such as comparing the similarity between different word pairs in vector space, assessing the performance of different models, and analyzing the relative degree of ethnic or gender bias in a corpus using different word lists.
\end{abstract}

\begin{figure*}
    \centering
    \includegraphics[width=0.8\linewidth]{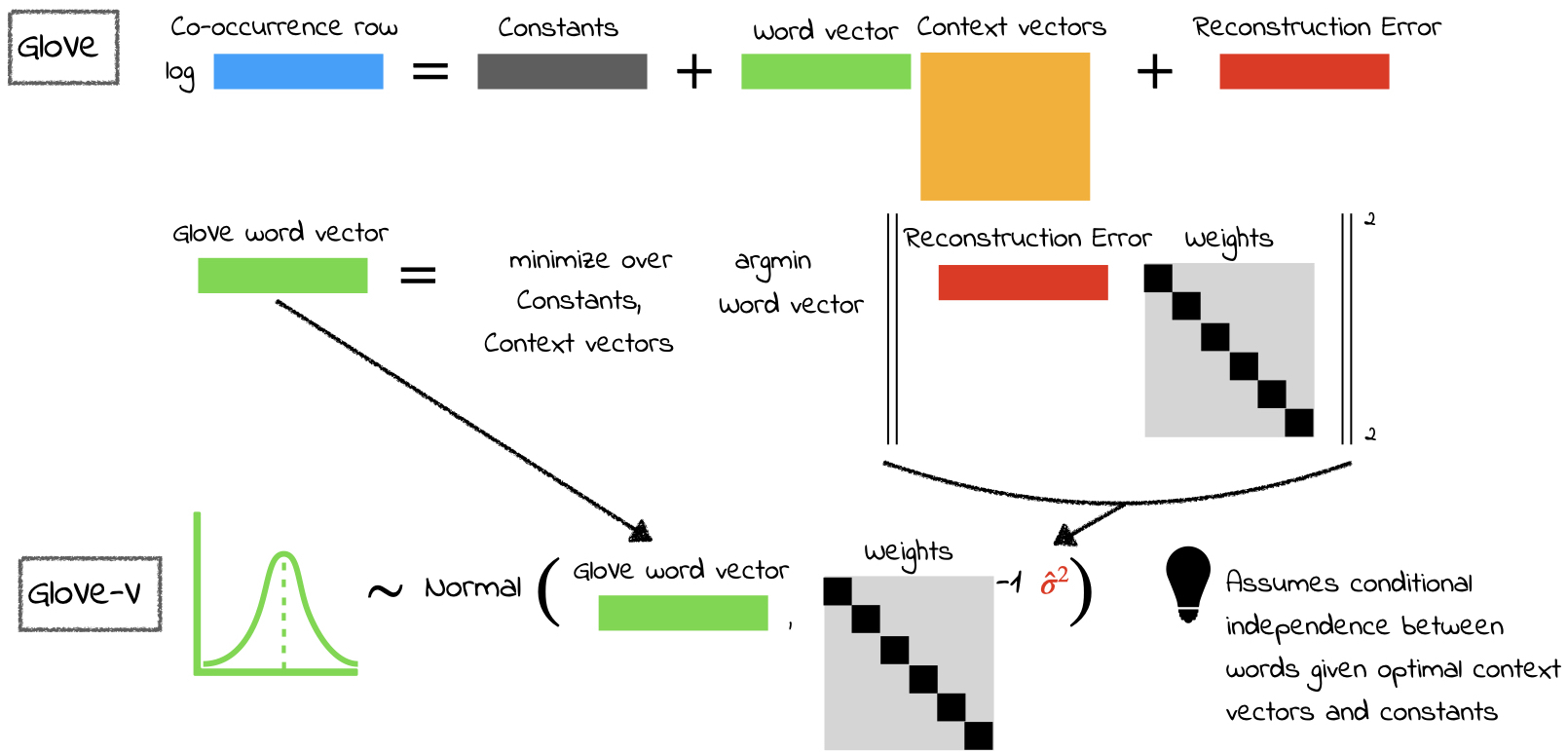}
    \caption{Conceptual diagram of the Glove-V method for one word. The top two rows illustrate the structural form and estimation of the original GloVe model \citep{pennington2014}, which models each row of a logged, weighted co-occurrence matrix as the product of a word vector and context vectors, plus constant terms. As shown in the third row, GloVe-V creates a distribution for the optimal GloVe word vector using the reconstruction error found through the GloVe minimization procedure. These distributions can be efficiently computed word-by-word by assuming conditional independence between words given the optimal context vectors and constants. } 
    \label{fig:diagram}
\end{figure*}

\section{Introduction}

Over the past decade, vector representations of words, or ``word embeddings,'' have become standard ways to quantify word meaning and semantic relationships due to their high performance on natural language tasks \citep{mikolov2013, pennington2014, levy2015}. Word embeddings are now ubiquitous in a wide variety of downstream computational social science applications, including charting the semantic evolution of words over time \citep{hamilton2016}, generating simplifications of scientific terminology \citep{kim2016}, comparing the information density of languages \citep{aceves2024}, assisting in legal interpretation \citep{choi2024}, and detecting societal biases in educational texts \citep{lucy2020}, historical corpora \citep{garg2018, charlesworth2022}, legal documents \citep{matthews2022,sevim2023}, political writing \citep{knoche_identifying_2019}, and annotator judgments \citep{davani2023}. Task performance metrics using word embeddings also factor prominently into the evaluation of more sophisticated, multimodal artificial intelligence systems, such as brain-computer interfaces \citep{tang2023} and adversarial text-to-image generation \citep{liu2023}.

Though a vast amount of research has relied on a relatively narrow set of word embedding models, no unified framework has emerged for representing statistical uncertainty in how accurately the word embeddings reconstruct the relationships implied by the sample of word co-occurrences. Intuitively, we should be less certain about a word's position in vector space the less data we have on its co-occurrences in the raw text \cite{ethayarajh2018, ethayarajh2019}. While it is generally standard practice in the social and natural sciences to check results for statistical significance, the vast majority of applications have relied exclusively on point estimates of embeddings, ignoring uncertainty even when training vectors over smaller corpora \citep[e.g.,][]{sevim2023,knoche_identifying_2019}. In select applications, approaches have ranged from a bootstrap on the \emph{documents} in the training corpus \citep{lucy2020}, to permutations of the \emph{word list} or lexicon  \citep{caliskan2017, garg2018}. 
While useful, such bootstrap and permutation approaches are computationally intractable on large datasets. Moreover, they address uncertainty from document or lexicon selection, even though embeddings are parameters of a data generating process on word co-occurrences, not on collections of documents or sets of words \citep{ethayarajh2019}. Until now, accounting for the fundamental uncertainty from the data-generating process has eluded the NLP community.

To fill this gap, we develop GloVe-V, a scalable, easy-to-use, computationally efficient method for approximating reconstruction error variances for the GloVe model \citep{pennington2014}, one of the most widely used word embedding models. 
Our approach leverages the core insight that if context vectors and constant terms are held fixed at optimal values, GloVe word embeddings are the optimal parameters for a multivariate normal probability model on a weighted log transformation of the rows of the co-occurrence matrix. If we assume that the rows of the co-occurrence matrix are independent given their context vectors and constant terms, the word embedding variances according to this likelihood are computationally tractable on large vocabularies. This assumption is reasonable and is also employed in other settings, such as measuring the influence of particular documents on downstream embedding statistics \citep{brunet2019}. Such variance estimates enable researchers to conduct rigorous assessments of model performance and principled statistical hypothesis tests on downstream tasks, responding to the need to account for statistical uncertainty and significance testing in natural language processing and machine learning \citep{card2020, dror2020, liao2021we,bowman_what_2021,ulmer_experimental_2022}.

Our contributions are threefold: (a)~we provide the statistical foundations for a principled notion of reconstruction error variances for GloVe word embeddings; (b)~we show that incorporating uncertainty can change conclusions about textual similarities, model selection, and textual bias; (c)~we provide a data release including pre-computed word embeddings and variances for the most frequently occurring words in the Corpus of American English (COHA), the largest corpus of historical American English that is widely used to track the usage and linguistic evolution of English terms over time \citep[e.g.,][]{ng2015, newberry2017, garg2018, xiao2023, charlesworth2024}.\footnote{These data products are intended for academic use. We also plan to release pre-computed word embeddings and variances for two larger corpora, Wikipedia \& Gigaword and DSIR Pile, to make our approach more readily accessible to researchers.} 

\section{Background on GloVe}
\label{sec:background}

We use upper case bold letters for matrices $\bs{X}$, lower case bold letters for vectors $\bs{x}$, and regular non-bolded letters for scalars $x$, except when indexing into a matrix or vector (i.e., the $ij^{th}$ entry of the matrix $\bs{X}$ is $\bs{X}_{ij}$). Sets are represented by script letters $\mathcal{X}$.

Word embedding models learn a shared vector space representation of words in a corpus. The training data are word co-occurrences in the corpus, which can be represented by a $V \times V$ co-occurrence matrix $\bs{X}$, where $\bs{X}_{ij}$ is the weighted number of times word $j$ appears in the context of word $i$,\footnote{Co-occurrence terms are usually weighted by the inverse of their distance from the center word.} and $V$ is the number of words in the vocabulary. A word embedding is a vector representation of a given word that emerges from the model. 

The GloVe word embedding model \citep{pennington2014} learns two embeddings $\bs{w}_k, \bs{v}_k \in \mathbb{R}^D$ for each word $k$, by minimizing the following cost function:
\begin{align}
\label{eq:cost}
    J = \sum_{i=1}^V \sum_{j = 1}^V f(\bs{X}_{ij})(\bs{w}_i^T\bs{v}_j + b_i + c_j - \log \bs{X}_{ij})^2
\end{align}
where $f(\bs{X}_{ij})$ is a non-negative weighting function with properties that ensure that very rare or very frequent co-occurrences do not receive too much weight, $b_i$ and $c_j$ are constant terms associated with word $i$ and $j$ respectively.\footnote{We follow the approach of the original authors and use $$f(x) = \begin{cases} (x/100)^{3/4} & \text{ if } $x < 100$ \\ 1 & \text{otherwise} \end{cases}$$} The vectors $\bs{v}_k$ are called ``context'' vectors and $\bs{w}_k$ are called ``center'' vectors, representing that word co-occurrences are defined based on words that appear within a fixed context window around a center word. The original implementation computed $\bs{w}_k + \bs{v}_k$ in a post-processing step to obtain a single embedding for a word $k$. In this paper, we focus on the center vector $\bs{w}_k$ as the embedding of interest for word~$k$.\footnote
  {While summing the context and center vectors can give useful performance gains, it is not always even a win. See the extensive discussion in \citep[secs.~3.3 and~6]{assylbekov2019,levy2015}.}

\section{Variance Derivation for GloVe-V}
\label{sec:method}

We now derive the GloVe-V variance estimator by recasting the optimization problem and recovering a probabilistic interpretation of GloVe embeddings.

\subsection{Reformulating the GloVe optimization problem}

The GloVe optimization problem using the cost in Equation~\ref{eq:cost} can be written in matrix form as 
\begin{align}
\label{eq:cost2}
\begin{split}
    &\min_{\bs{b}, \bs{c}, \bs{W}, \bs{V}} \|\bs{F} \odot \bs{R}\|_F  \\
    &s.t. \quad \bs{R} = \log\bs{X} - \bs{W}^T\bs{V} - \bs{b}\bs{1}^T - \bs{1}\bs{c}^T \\
    &\text{rank}(\bs{W}), \text{rank}(\bs{V}) \leq D 
\end{split}
\end{align}
where $\bs{F}_{ij} = \sqrt{f(\bs{X}_{ij})}$, $\bs{w}_i$ and $\bs{v}_i$ are the $i^{th}$ and $j^{th}$ columns of matrices $\bs{W}$ and $\bs{V}$ respectively, $b_i$ and $c_j$ are the $i^{th}$ and $j^{th}$ elements of vectors $\bs{b}$ and $\bs{c}$ respectively, and $\odot$ is the element-wise product. Equation~\ref{eq:cost2} is an element-wise  weighted low-rank approximation problem that can be solved in two steps \citep[e.g.,][]{markovsky2012}:
\begin{align}
    \min_{\bs{b} \in \mathbb{R}^V, \bs{c} \in \mathbb{R}^V, \bs{V} \in \mathbb{R}^{V \times D}} \min_{\bs{W} \in \mathbb{R}^{V \times D}} \|\bs{F} \odot \bs{R}\|_F  
\end{align}
That is, holding the choice of $(\bs{b}, \bs{c}, \bs{V})$ fixed at their globally optimal values $(\bs{b}^*, \bs{c}^*, \bs{V}^*)$, the inner minimization to find the optimal $\bs{W}$ decomposes to $V$ weighted least squares projections with solutions:
\begin{align}
\label{eq:wstar}
    \bs{w}^*_i = (\bs{V}_{\mathcal{K}}^{*T}\bs{D}_{\mathcal{K}}\bs{V}^*_{\mathcal{K}})^{-1}\bs{V}_{\mathcal{K}}^{*T}\bs{D}_{\mathcal{K}}(\log\bs{x}_i - b^*_i\bs{1} - \bs{c}^* )
\end{align}
where $\mathcal{K}$ is the set of column indices with non-zero co-occurrences for word $i$, $\bs{V}_{\mathcal{K}}^*$ is a matrix whose columns belong to the set $\{\bs{v}^*_j: j \in \mathcal{K}\}$, $\bs{D}_{\mathcal{K}} = \text{diag}(\{\bs{F}_{ij}^2 : j \in \mathcal{K}\})$, and $\bs{x}_i$ is the $i^{th}$ row of $\bs{X}$.

\subsection{A probabilistic model for an approximate problem}

Recasting the optimization problem in this fashion allows us to state the natural probabilistic model for which $\bs{w}_i^*$ in Equation~\ref{eq:wstar} is optimal. Conditional on the optimal context vector subspace spanned by $\bs{V}^*$ and optimal constant vectors $(\bs{b}^*, \bs{c}^*)$, we write the following weighted multivariate normal model for the rows of $\bs{X}$:
\begin{align}
\label{eq:prob}
\begin{split}
&\log\bs{x}_{i} = b^*_i\bs{1} +  \bs{c}^* + \bs{w}_i^{T}\bs{V}_{\mathcal{K}}^* + \bs{e}_{i}
\\
&\bs{e}_{i} \sim \mathcal{N}(\bs{0}, \bs{D}_{\mathcal{K}}^{-1}\sigma^2_i)
\end{split}
\end{align}
where we always have $(\bs{D}_{\mathcal{K}})_{ii} > 0$ due to the fact the $\bs{X}_{ij} > 0$ for $j \in \mathcal{K}$, and we assume that the rows of $\log\bs{X}$ are independent given the optimal parameters, i.e., $\log\bs{x}_{i} | \bs{b}^*, \bs{c}^*, \bs{V}_{\mathcal{K}}^* \perp \log\bs{x}_{j} |  \bs{b}^*, \bs{c}^*, \bs{V}_{\mathcal{K}}^*  $ for $i \not = j$.\footnote{\citet{brunet2019} rely on the same assumption to make their method for approximating the influence of a group of documents on word association statistics computationally tractable.} Then, under standard assumptions for weighted least squares estimators \citep[e.g.,][]{romano2017}, the covariance matrix for $\bs{W}$ simplifies into a $VD \times VD$ block diagonal matrix with the $i^{th}$ $D \times D$ block given by: 
\begin{align}
\label{eq:sigma}
    \bs{\Sigma}_i = \sigma_i^2\left(\sum_{j \in \mathcal{K}} f(\bs{X}_{ij})\bs{v}^*_j(\bs{v}_j^*)^{T} \right)^{-1} 
\end{align}
We can then estimate $\sigma_i^2$, the reconstruction error for word $i$, with the plug-in estimator:
\begin{align}
\begin{split}
 &\hat{\sigma}_i^2 = \\
 &\frac{1}{|\mathcal{K}| - D} \sum_{j \in \mathcal{K}}f(\bs{X}_{ij})(\log\bs{X}_{ij} - b^{*}_i - c^{*}_j - \bs{w}_i^{*T}\bs{v}^{*}_j)^2
\end{split}
\end{align}

\begin{figure}
    \centering
    \includegraphics[width=0.9\columnwidth]{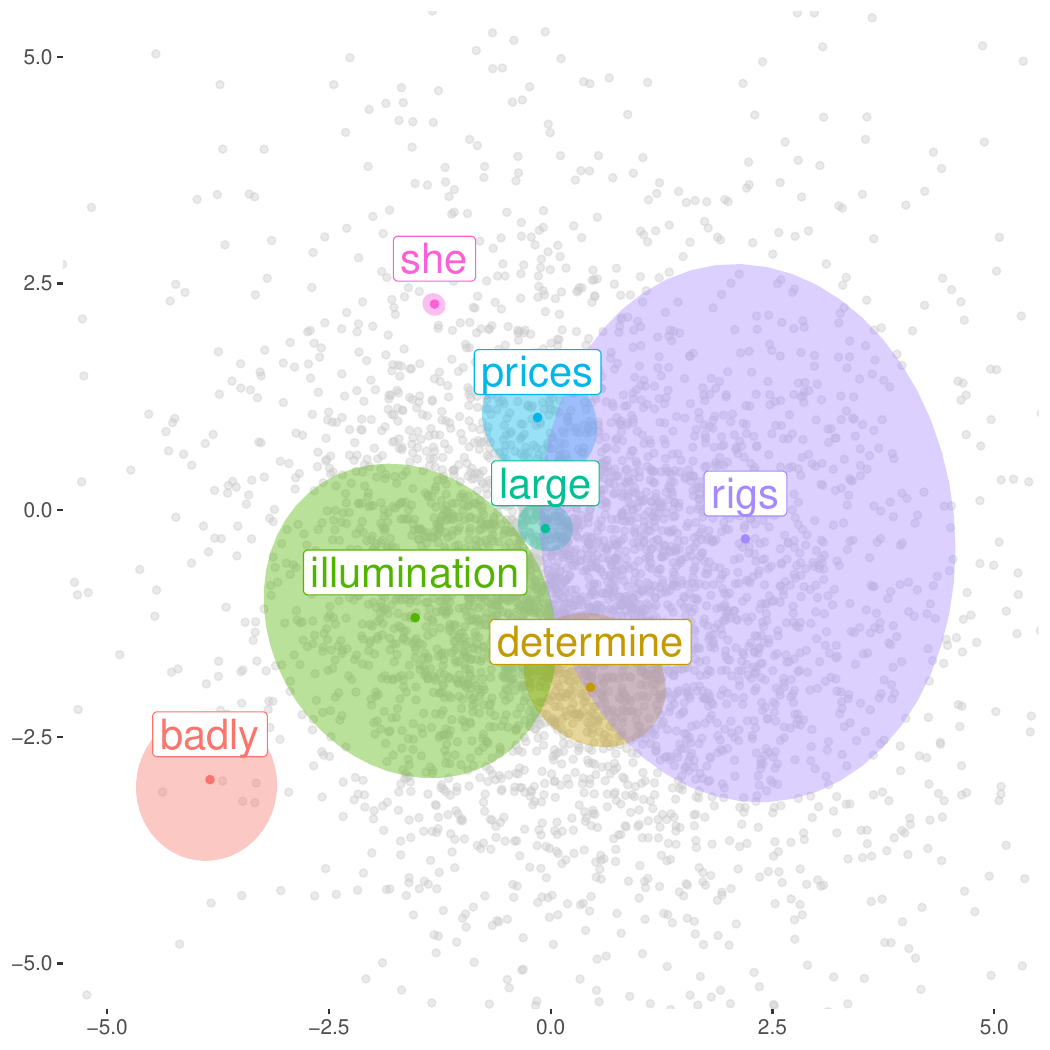}
    \caption{\textbf{Uncertainty in word embedding locations.} Two-dimensional representations of GloVe word embeddings trained on COHA (1900--1999), along with ellipses drawn around 100 draws from the estimated multivariate normal distribution from Equation~\ref{eq:prob} for a random subset of words. Lower frequency words like ``rigs'' and ``illumination'' have more uncertainty in their estimated positions in the vector space than high frequency words like ``she'' and ``large.'' }
    \label{fig:vector2d}
\end{figure}

\subsection{Estimation}\label{sec:practical_estimation}

The covariance estimator in Equation~\ref{eq:sigma} is only valid for words that co-occur with a greater number of unique context words $|\mathcal{K}|$ than the dimensionality of the vectors $D$. A simple way to increase the coverage of the variances in smaller corpora is to reduce the dimensionality of the word vectors. However, when $|\mathcal{K}| \approx D$, numerical problems with computing the inverse in Equation~\ref{eq:sigma} are likely to occur even if it is technically possible to compute an inverse. To address these numerical issues, for each word whose Hessian block $\bs{H}_i = \sum_{j \in \mathcal{K}} f(\bs{X}_{ij})\bs{v}^*_j(\bs{v}_j^*)^{T}$ has a condition number that implies numerical error in excess of 1e-10 in its inverse, we instead compute the Moore-Penrose pseudo-inverse of $\bs{H}_i$ as $\bs{V}\bs{\Lambda}^{+}\bs{U}^T$ \citep{golub2013}, where $\bs{H}_i = \bs{U}\bs{\Lambda}\bs{V}^T$ is the singular value decomposition of $\bs{H}_i$ and $\bs{\Lambda}^{+}_{jj} = 1/\bs{\Lambda}_{jj}$ if $\bs{\Lambda}_{jj} > \text{1e-3} \times \max_j \bs{\Lambda}_{jj}$ and $0$ otherwise. This technique effectively drops dimensions that are predominantly noise in the Hessian block when computing the inverse.

\subsection{Propagating uncertainty}

With this derivation in hand, propagating variance to downstream tasks is straightforward. For differentiable test statistics, such as the cosine similarity between two word embeddings, the most computationally efficient approach is to use the delta method for asymptotic variances \citep{van2000}. Using a first-order Taylor series approximation to the test statistic, the delta method states that if $\sqrt{n}(\bs{W} - \hat{\bs{W}})$ converges to $\mathcal{N}(\bs{0}, \bs{\Sigma})$, then $\sqrt{n}(\phi(\bs{W}) - \phi(\hat{\bs{W}}))$ converges to $\mathcal{N}(\bs{0}, \phi^{\prime}(\bs{W})^T\bs{\Sigma}\phi^{\prime}(\bs{W}))$, where $\phi(\cdot)$ is a differentiable function of $\bs{W}$ and $\phi^{\prime}(\cdot)$ is its gradient with respect to $\bs{W}$. If the test statistic only depends on a subset of words in the vocabulary, the computation is quite efficient due to the fact that the gradient will be sparse. 

For a broader class of test statistics, researchers can  repeatedly draw from the estimated multivariate normal distribution in Equation~\ref{eq:sigma} and recalculate the test statistic of interest \citep{tanner_tools_1996}, which is much more computationally efficient than a bootstrap on the full embedding model. In our code repository (\url{github.com/reglab/glove-v}), 
we provide a tutorial and starter code to apply the GloVe-V framework to any downstream test statistic using this method.

\begin{figure}
    \centering
    \includegraphics[width=\columnwidth]{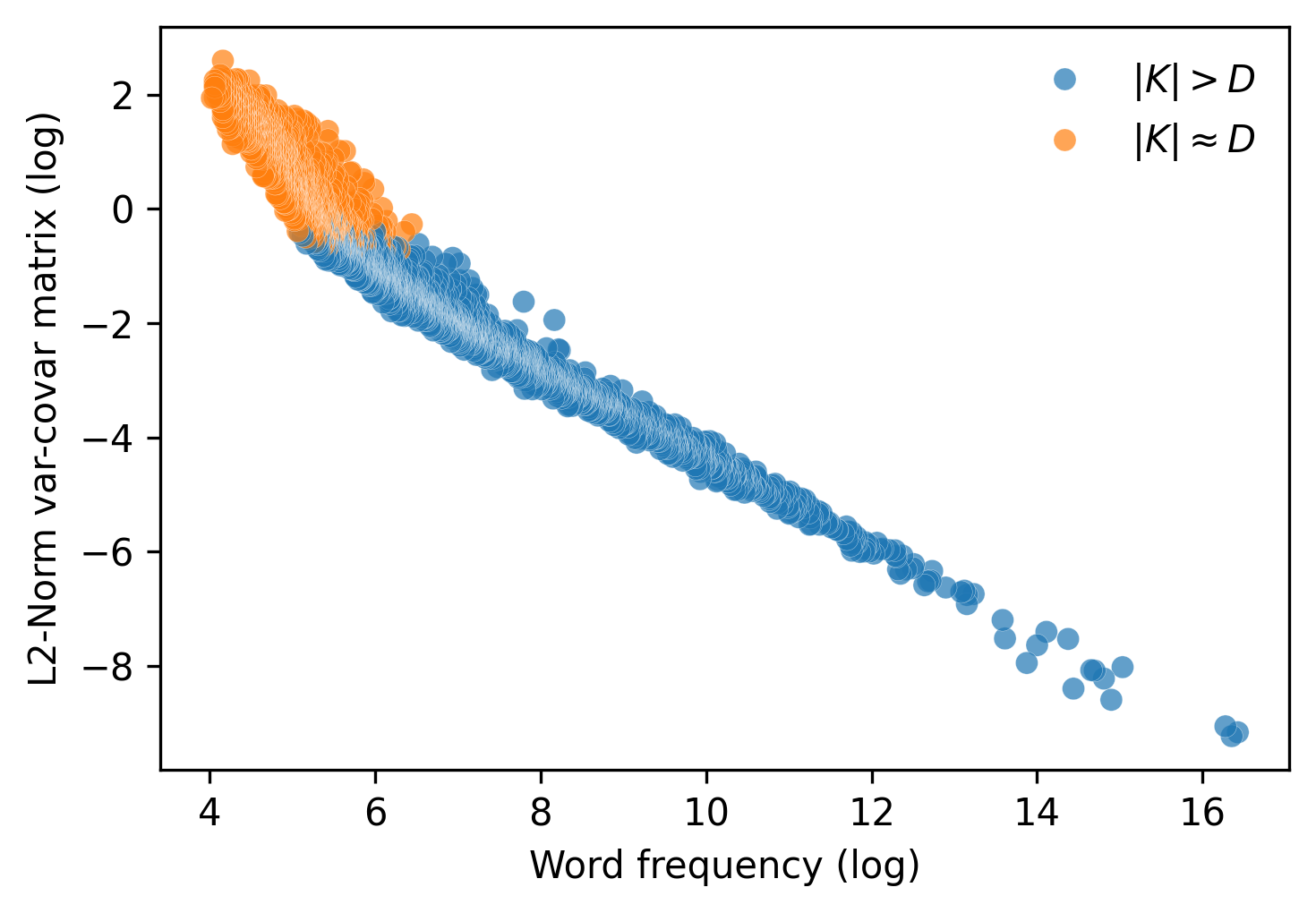}
    \caption{\textbf{Word-level relationship between GloVe-V variances and frequency on COHA (1900--1999).} L2-norm of the diagonal of $\hat{\bs{\Sigma}}$ from Equation~\ref{eq:sigma} ($x$-axis, on a  $\log_{10}$ scale) plotted against logged word frequencies ($y$-axis, on a $\log_{10}$ scale) for a subset of 5,000 words randomly sampled in proportion to word frequency. The variances for words colored in orange are computed as discussed in Section \ref{sec:practical_estimation}. }
    \label{fig:validationl2}
\end{figure}

\section{Results}\label{sec:results}

\begin{figure}[ht]
    \centering
    \includegraphics[width=0.95\columnwidth]{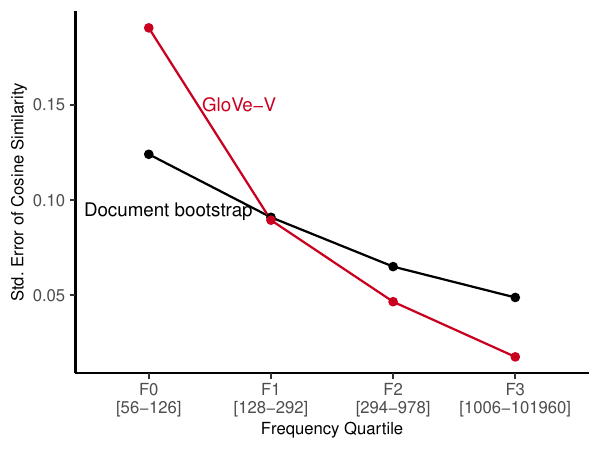}
    \caption{\textbf{Comparison between document bootstrap and GloVe-V standard errors for cosine similarity.} The average standard error of the cosine similarity between 1,600 randomly sampled word pairs ($y$-axis) as a function of the frequency for the word pair ($x$-axis with word frequency ranges in brackets), using the document bootstrap approach and Glove-V using the delta method. The GloVe-V standard errors are more sensitive to word frequency and are more efficient to compute.
      }
    \label{fig:bootstrap}
\end{figure}

To build intuition and demonstrate the usefulness of GloVe-V variances, we now provide empirical results using the Corpus of Historical American English (COHA) for the $20^{th}$ century, which contains English-language texts from a balanced set of genres (fiction, non-fiction, magazines, and newspapers) from 1900--1999 \citep{davies2012}.\footnote{The data were accessed under a standard academic license in accordance with the data usage restrictions at \url{https://www.corpusdata.org/restrictions.asp}.} For all examples, we use 300-dimensional GloVe embeddings using a symmetric context window of 8 words.\footnote{See Appendix \ref{app:training} for more details on model training.} The empirical examples show how GloVe-V variances can move researchers in NLP towards emerging best practices for incorporating hypothesis testing in natural language tasks and downstream analyses \citep{card2020}. 

\subsection{Uncertainty in word embedding locations}

We first show that the variances can represent reconstruction error uncertainty in the locations of individual words in vector space due to data sparsity. Figure~\ref{fig:vector2d} plots a two-dimensional representation of the word embeddings, with ellipses drawn around 100 draws from the estimated multivariate normal distribution in Equation~\ref{eq:prob} for a random subset of words. The size of the ellipses reflects the fact that, based on the underlying co-occurrence matrix, we are more certain about the positions of higher frequency words like ``she'' and ``large'' than lower frequency words like ``illumination'' and ``rigs.'' The higher uncertainty for lower frequency words is a structural feature of the estimated covariance matrices themselves. \figurename~\ref{fig:validationl2} demonstrates this feature by plotting the word-level frequency ($x$-axis, on a $\log_{10}$ scale) against the L2-norm of the diagonal of the estimated $\hat{\bs{\Sigma}}$ from Equation~\ref{eq:sigma} ($y$-axis, on a $\log_{10}$ scale) for a random subset of words (sampled in proportion to their frequency). The magnitude of the variances for the word embedding parameters decreases smoothly as the word frequency increases. Where $|\mathcal{K}| \approx D$ (highlighted in orange in Figure~\ref{fig:validationl2}), the estimation approach described in Section~\ref{sec:practical_estimation} provides a reasonable estimate of the variance.

\subsection{Comparison to document bootstrap}

We now provide intuition for why the GloVe-V variances may in many instances be preferable to the document bootstrap for hypothesis testing on downstream test statistics. The document bootstrap is a computationally intensive approach to capturing word embedding instability that repeatedly resamples documents from the corpus, recomputes the word embeddings, and recalculates the test statistic of interest \citep{antoniak2018}. To re-purpose this approach in order to conduct a hypothesis test, we must subscribe to the uncertainty framework that the corpus itself is randomly sampled from a hypothetical population of documents. However, as described in Section~\ref{sec:method}, this framework does not match the statistical micro-foundations under which the embeddings themselves are estimated, causing a mismatch between the notion of document-level uncertainty and the estimation target of the embeddings.

Figure~\ref{fig:bootstrap} shows that document-level uncertainty can either underestimate or overestimate the variance of a downstream test statistic compared to the reconstruction error uncertainty given by GloVe-V, depending on the distribution of words across documents. A word that is used infrequently but in the same way across many documents may have low document-level uncertainty because each bootstrap sample will yield similar (but sparse) co-occurrence counts for that word, even if the reconstruction error remains high for each bootstrapped estimate.  Conversely, a word that is extremely common in only a few documents may have high document-level uncertainty because many bootstrap samples will drop the majority of documents containing that word, even if the reconstruction error is low when all documents are included. Rather than choose one over the other, researchers can use each method for different purposes: tools like the document bootstrap or more computationally efficient analogs \citep[e.g.,][]{brunet2019} can be used to assess sensitivity of results to particular documents, while GloVe-V can be used to conduct hypothesis tests under a coherent statistical framework, holding the corpus fixed.

\section{GloVe-V enables principled significance testing}\label{sec:app}

We now show how GloVe-V enables statistical significance  testing, addressing the increasing recognition for NLP to move beyond point estimates alone \citep{card2020, liao2021we}. GloVe-V can also help researchers assess when a corpus is underpowered for specific inferences, as we illustrate below.   

\subsection{Uncertainty in $k$ nearest neighbors}

\begin{figure}[t]
    \centering
    \includegraphics[ width=0.8\columnwidth]{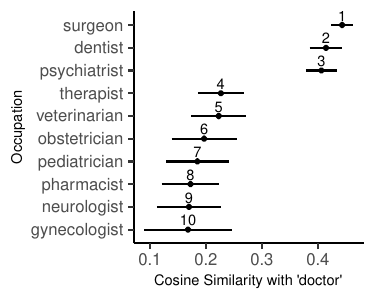}
    \caption{\textbf{Nearest neighbors with uncertainty.} Healthcare occupations ($y$-axis) ranked by their cosine similarity with ``doctor'' ($x$-axis), with the nearest neighbor ranking based on the point estimate above each point, and $95\%$ GloVe-V uncertainty intervals.   }
    \label{fig:occupations}
\end{figure}

Word similarity, including $k$ nearest neighbor lists, informs performance evaluation for both embedding models and more sophisticated artificial intelligence systems \citep[e.g.][]{mikolov_efficient_2013,levy_linguistic_2014,linzen_issues_2016, borah2021, tang2023, liu2023}.
Using only point estimates to evaluate word similarity, however, leaves the researcher with no sense of which word similarities are inherently less certain because they are based on less co-occurrence data in the underlying corpus. Uncertainty in neighbor rankings is particularly consequential for word similarity tasks, which depend on the ranking of different word pairs, and for word analogy tasks, which are typically solved by finding nearest neighbors in the embedding space.\footnote{In the case of analogy tasks, the solution typically relies on the nearest neighbor to some linear transformation of the word embeddings belonging to the words in the analogy~\citep{levy_linguistic_2014}, rather than the nearest neighbor to a specific word as illustrated in our example. 
} As an example of this dilemma, \figurename~\ref{fig:occupations} plots the cosine similarity between ``doctor'' and a list of healthcare occupation words along with GloVe-V uncertainty intervals. Based on the point estimates, we can assign a nearest-neighbor rank to each word by their proximity to ``doctor'' (printed above the point estimate in Figure~\ref{fig:occupations}). However, for the top three neighbors (and between neighbours 4 through 10), we cannot statistically distinguish the ranks (e.g., $p = 0.10$ for the difference in the cosine similarity of ``doctor" and ``surgeon" relative to that of ``doctor" and ``dentist"). Which neighbor is the ``nearest'' is therefore subject to considerable uncertainty that would be invisible without incorporation of the GloVe-V variances.

\subsection{Uncertainty in model performance}

\begin{figure*}[ht]
    \centering
    \includegraphics[ width=0.95\textwidth]{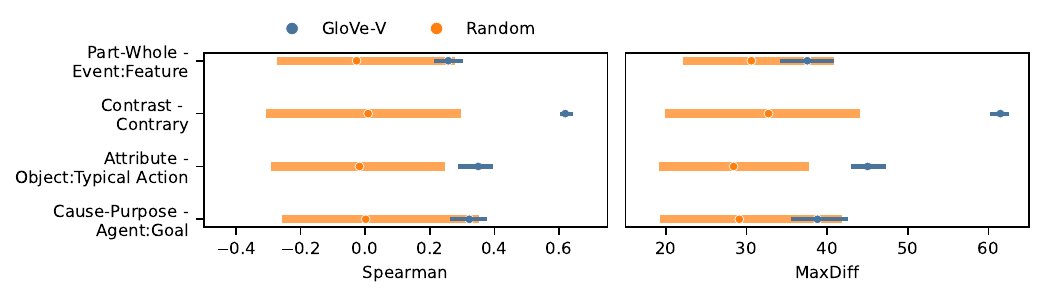}
    \caption{\textbf{Accounting for uncertainty in word embedding performance assessments using the SemEval-2012 Task 2 of \citet{jurgens_semeval-2012_2012} on COHA 1900--1999.} The task measures the degree of relational similarity of word pairs using the Spearman correlation and the MaxDiff choice procedure on a taxonomy that comprises 79 types of relations across 10 different classes (\textit{e.g.}, contrast, part-whole, cause-purpose). We present GloVe-V results on a subset of relations, along with a Random baseline that randomly rates the word pairs in each relation.}
    \label{fig:closed_analogy}
\end{figure*}

Performance on analogy tasks is a canonical approach to word embedding model evaluation \citep{mikolov2013,pennington2014,levy2015}. Closed-list relational similarity tasks such as SemEval-2012 \citep{jurgens_semeval-2012_2012} are structured so that models can be benchmarked against random baselines, in which pairs of words are randomly related to each other to establish a lower bound on expected performance. Figure~\ref{fig:closed_analogy} presents the performance of GloVe compared to a random benchmark using two evaluation metrics on four relational similarity tasks \citep[see][for details on the tasks and metrics]{jurgens_semeval-2012_2012}. While the point estimates for performance suggest that GloVe outperforms the random baseline on all tasks, adding uncertainty to both point estimates reveals that we can only claim significantly higher performance than random on two of the four relations ($p =0.08$, $p  < 0.001$, $p = 0.02$, $p = 0.08$ for relations 1--4, respectively, for the MaxDiff metric). The GloVe-V intervals also allow us to distinguish the performance of GloVe across different relational similarity tasks. While we can say that GloVe performs better on ``Contrast'' vs.\ ``Cause-Purpose'' ($p = 0.03$ for the MaxDiff metric), we cannot claim that it does better on ``Cause-Purpose'' than ``Part-Whole'' ($p=0.92$), even though the point estimates suggest better performance on the former task.\footnote{The relations contain about about 27 to 45 word pairs: 1) \textit{Contrast - Contrary}: pairs of opposite words (e.g., ``dull" and ``bright"); 2) \textit{Part-Whole - Event: Feature}: pairs in which one word is a part of the event given by the first word (e.g., ``carnival" and ``rides"); 3) \textit{Attribute - Object: Typical Action}: pairs in which one word is a characteristic action of the other (e.g., ``heart" and ``beat"); 4) \textit{Cause-Purpose - Agent: Goal}: pairs in which one word is a typical objective of the agent given by the first word (e.g., ``painter" and ``portrait").} 

\subsection{Uncertainty in word embedding bias}\label{sec:bias}

\begin{figure*}[ht]
    \centering
    \includegraphics[ width=0.48\textwidth]{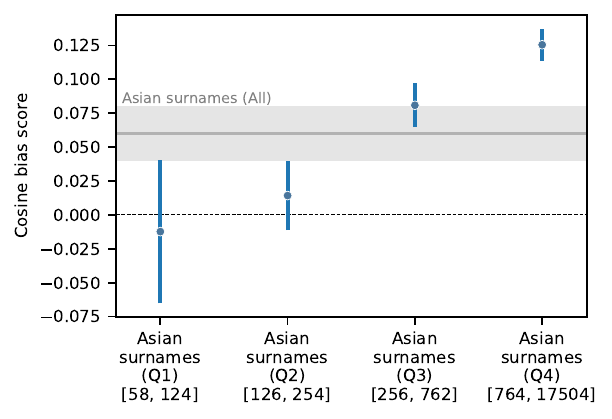}
    \includegraphics[ width=0.48\textwidth]{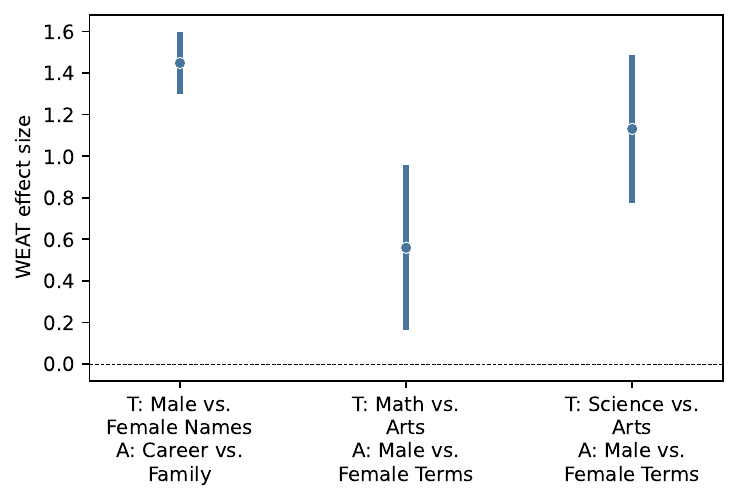}
    \caption{\textbf{Ethnicity and gender bias scores.} 
    \textbf{a)} Average Asian bias scores with GloVe-V uncertainty using the cosine bias score of \citet{garg2018} in COHA 1900--1999 for different Asian surname lists. The gray line and  shaded gray area represent the point estimate and 95\% GloVe-V uncertainty interval, respectively, of the bias score on a master list of Asian surnames. The points and error bars in blue represent the bias score computed on different subsets of the list, grouped according to the number of times they appear in the corpus.
    \textbf{b)} Gender bias scores for three types of bias tests using the WEAT effect size of \citet{caliskan2017} with GloVe-V uncertainty. } 
    \label{fig:bias_queries}
\end{figure*}

Measurement of societal biases in text is a popular downstream application using word embeddings \citep[e.g.,][]{garg2018, lucy2020, charlesworth2022, matthews2022, sevim2023}. These types of studies compare similarity between curated sets of words to test the prevalence of societal biases in text. For example, if a set of female-oriented words is closer to a set of family-oriented words than to career-oriented words, and this relationship is stronger relative to the same comparison using male-oriented words, that is evidence of a gender bias \citep{bolukbasi2016, caliskan2017}. To represent uncertainty in these comparisons, researchers typically use a permutation test or bootstrap on the words included in each set \citep[e.g.,][]{caliskan2017, garg2018}, but others have noted that these types of uncertainty measures (which account for uncertainty in word selection) are not designed to account for the sparsity of word co-occurrences that form the basis for the comparisons \citep{ethayarajh2019}. 

It is especially important to account for uncertainty due to sparsity in applications where the analysis relies on infrequently occurring words such as surnames, which are often used to measure demographic bias \citep[e.g.,][]{caliskan2017, garg2018, swinger2019}. Researchers typically drop lower frequency surnames altogether from their analyses \citep[e.g.,][]{garg2018} because they have no way of representing the higher uncertainty in the embedded positions for lower frequency surnames using only point estimates. But such curation runs the risk of sacrificing the representativeness of the word lists involved and the generalizability of the conclusions \citep{antoniak2021}. Using a measure of anti-Asian bias in the COHA corpus based on \citet{garg2018}, the left panel of Figure~\ref{fig:bias_queries} shows how GloVe-V variances can automatically provide information on co-occurrence sparsity for researchers.\footnote{Appendix~\ref{app:delta} provides delta method derivations for the bias statistics used in Figure~\ref{fig:bias_queries}.} The bias measure on the $y$-axis computes the average cosine similarity between a set of Asian surnames and a set of 20 Otherization words,\footnote{This word list is primarily composed of adjectives used to describe people as outsiders, such as \textit{monstruous}, \textit{devious}, and \textit{bizarre}.} relative to a set of White surnames. A more positive bias score indicates that Asian surnames are more closely related to these negative Otherization words compared to White surnames. 

Figure~\ref{fig:bias_queries} shows that the anti-Asian bias estimate in COHA becomes more positive for more frequently appearing surnames, such that relying only on the most frequent surnames produces an exaggerated result relative to the full set of Asian surnames.\footnote{See Appendix \ref{app:surname} for details on how we compiled this surname list.} This is likely due to the fact that the most frequently occurring surnames tend to be from historical figures such as ``Ghandi,'' ``Mao,'' and ``Mohammed,'' which are clearly not representative of the entire class of Asian surnames. Using GloVe-V, low and high frequency words can be seamlessly combined into a single bias interval that represents the combined uncertainty in all the estimated word positions (shown as a gray interval on the plot), without having to drop any surnames and sacrifice generalizability.\footnote{GloVe-V intervals can also help researchers determine when a bias time trend can be supported by the co-occurrence data. Estimating the anti-Asian bias separately for each decade of COHA with GloVe-V uncertainty intervals, for example, reveals no significant trend in the bias score over time, suggesting that time trends such as those reported in \citet{garg2018} should also be checked for statistical significance.}

GloVe-V intervals can also be useful for studying high frequency word lists because they allow researchers to make statistical comparisons between \emph{types} of bias. The right panel of Figure~\ref{fig:bias_queries} provides an example of three gender bias queries with GloVe-V intervals for the Word Embedding Association Test (WEAT) effect size, a cosine-similarity-based test \citep{caliskan2017}: (a) male vs.\ female names and words related to \textit{career} vs.\ \textit{family}; (b) male vs.\ female terms and words related to \textit{math} vs.\ \textit{arts}; and (c) male vs.\ female terms and words related to \textit{science} vs.\ \textit{arts}. While the point estimate for type (a) is higher than those for both (b) and (c), the GloVe-V intervals allow us to reject the null that (a) is equal to (b) ($p < 0.001$), but not the null that (a) is equal to (c) ($p = 0.11$). In this case, the GloVe-V intervals guard against making unsubstantiated claims about which types of bias are strongest.

\section{Conclusion}

In this paper, we have derived, computed, and demonstrated the utility of GloVe-V, a new method to represent uncertainty in word embedding locations using the seminal GloVe model. The GloVe-V variances provide researchers with the ability to easily propagate uncertainty in the word embeddings to downstream test statistics of interest, such as word similarity metrics that are used in both evaluation and analyses involving word embeddings. Unlike methods such as the document and word list bootstrap, the method is computationally efficient even on large corpora and represents uncertainty due to sparsity in the underlying co-occurrence matrix, which is often invisible in downstream analyses that use only the estimated word embeddings. 

As we have shown in Section~\ref{sec:results}, incorporating uncertainty into downstream analyses can have consequential impacts on the conclusions researchers draw, and should be a best practice moving forward for studies that use word embeddings to infer semantic meaning. Finally, while outside the scope of the current study, we note that the contextual word and passage representations of transformer large language models are also point estimates and similar questions of embedding uncertainty apply when using such models as well.

\section*{Limitations}

While useful in many applications, the GloVe-V method comes with certain limitations. First, the variances can only be computed for words whose number of context words exceeds the embedding dimensionality. This limitation can easily be minimized by reducing the dimensionality of the vectors for small corpora; for example, variances can be computed for $96\%$ of the word embeddings in the relatively small \emph{New York Times} Annotated Corpus (NYT) with 50-dimensional vectors, compared to $36\%$ of embeddings with 300-dimensional vectors. Second, researchers need access to the co-occurrence matrix if they wish to compute the variances themselves, since it relies on an empirical estimate of the reconstruction error. Third, the methodology in this paper applies solely to the GloVe embedding model because of its statistical foundations. That said, this model is one of the most-cited word embedding models in current use and has been shown to have better stability and more intuitive geometric properties than competing models \citep{mimno2017, wendlandt2018}.

Finally, the uncertainty captured by GloVe-V intervals is due to sparsity in the underlying co-occurrence matrix, which is only one of many types of uncertainty one could consider in embedded locations for words. Other types of uncertainty that are held fixed in GloVe-V include instability due to the documents included in the corpus \citep[e.g.,][]{antoniak2018}, uncertainty due to the hyper-parameters of the model \citep[e.g.,][]{borah2021}, and statistical uncertainty in the estimated context vector positions and bias terms, which are treated as constants in the variance computation for computational tractability. Along with the conditional independence assumption on words, treating these terms as constants is necessary to reduce the number of free parameters in the model and allow a tractable variance computation. These sorts of independence assumptions are becoming standard practice to enable computational efficiency for models with a large number of parameters -- the same assumption, for example, has been successfully employed to develop a computationally efficient approximation to a document bootstrap \citep{brunet2019}.

Despite these limitations, we found by trying a number of other approaches (detailed in Appendix~\ref{app:alt}) that GloVe-V strikes a desirable balance between maintaining the model's probabilistic foundations for enhanced statistical rigor, and preserving computational tractability for practical purposes.  

\section*{Acknowledgements}
We thank Rishi Bommasani, Matthew Dahl, Neel Guha, Peter Henderson, Varun Magesh, Joel Niklaus, Derek Ouyang, Dilara Soylu, Faiz Surani, Mirac Suzgun, Lucia Zheng, and participants at the 2024 Stanford Data Science Conference for helpful comments and discussions. 

\bibliography{custom}

\begin{thebibliography}{54}
\providecommand{\natexlab}[1]{#1}

\bibitem[{Aceves and Evans(2024)}]{aceves2024}
Pedro Aceves and James~A Evans. 2024.
\newblock \href {https://doi.org/10.1038/s41562-024-01815-w} {Human languages
  with greater information density have higher communication speed but lower
  conversation breadth}.
\newblock \emph{Nature Human Behaviour}, 8:1--13.

\bibitem[{Antoniak and Mimno(2018)}]{antoniak2018}
Maria Antoniak and David Mimno. 2018.
\newblock \href {https://doi.org/10.1162/tacl_a_00008} {Evaluating the
  stability of embedding-based word similarities}.
\newblock \emph{Transactions of the Association for Computational Linguistics},
  6:107--119.

\bibitem[{Antoniak and Mimno(2021)}]{antoniak2021}
Maria Antoniak and David Mimno. 2021.
\newblock \href {https://doi.org/10.18653/v1/2021.acl-long.148} {Bad seeds:
  Evaluating lexical methods for bias measurement}.
\newblock In \emph{Proceedings of the 59th Annual Meeting of the Association
  for Computational Linguistics and the 11th International Joint Conference on
  Natural Language Processing (Volume 1: Long Papers)}, pages 1889--1904.

\bibitem[{Assylbekov and Takhanov(2019)}]{assylbekov2019}
Zhenisbek Assylbekov and Rustem Takhanov. 2019.
\newblock \href {https://doi.org/10.1613/jair.1.11368} {Context vectors are
  reflections of word vectors in half the dimensions}.
\newblock \emph{Journal of Artificial Intelligence Research}, 66:225--242.

\bibitem[{Blundell et~al.(2015)Blundell, Cornebise, Kavukcuoglu, and
  Wierstra}]{blundell_weight_2015}
Charles Blundell, Julien Cornebise, Koray Kavukcuoglu, and Daan Wierstra. 2015.
\newblock \href {https://proceedings.mlr.press/v37/blundell15.html} {Weight
  uncertainty in neural network}.
\newblock In \emph{International conference on machine learning}, pages
  1613--1622. PMLR.

\bibitem[{Bolukbasi et~al.(2016)Bolukbasi, Chang, Zou, Saligrama, and
  Kalai}]{bolukbasi2016}
Tolga Bolukbasi, Kai-Wei Chang, James~Y Zou, Venkatesh Saligrama, and Adam~T
  Kalai. 2016.
\newblock \href
  {https://proceedings.neurips.cc/paper_files/paper/2016/hash/a486cd07e4ac3d270571622f4f316ec5-Abstract.html}
  {Man is to computer programmer as woman is to homemaker? debiasing word
  embeddings}.
\newblock In \emph{Advances in Neural Information Processing Systems},
  volume~29. Curran Associates, Inc.

\bibitem[{Book(2020)}]{book_training_2020}
Jonathan~W. Book. 2020.
\newblock \href
  {https://lup.lub.lu.se/luur/download?func=downloadFile&recordOId=9008040&fileOId=9008041}
  {Training bayesian neural networks: A study of improvements to training
  algorithms}.
\newblock Master thesis.

\bibitem[{Borah et~al.(2021)Borah, Barman, and Awekar}]{borah2021}
Angana Borah, Manash~Pratim Barman, and Amit Awekar. 2021.
\newblock \href {https://doi.org/10.1145/3465336.3475098} {Are word embedding
  methods stable and should we care about it?}
\newblock In \emph{Proceedings of the 32nd ACM Conference on Hypertext and
  social media}, pages 45--55.

\bibitem[{Bowman and Dahl(2021)}]{bowman_what_2021}
Samuel~R. Bowman and George~E. Dahl. 2021.
\newblock \href {https://aclanthology.org/2021.naacl-main.385.pdf} {What will
  it take to fix benchmarking in natural language understanding?}
\newblock In \emph{Proceedings of the 2021 Conference of the North American
  Chapter of the Association for Computational Linguistics: Human Language
  Technologies}, page 4843–4855.

\bibitem[{Brunet et~al.(2019)Brunet, Alkalay-Houlihan, Anderson, and
  Zemel}]{brunet2019}
Marc-Etienne Brunet, Colleen Alkalay-Houlihan, Ashton Anderson, and Richard
  Zemel. 2019.
\newblock \href {https://proceedings.mlr.press/v97/brunet19a.html}
  {Understanding the origins of bias in word embeddings}.
\newblock In \emph{International conference on machine learning}, pages
  803--811. PMLR.

\bibitem[{Bruni et~al.(2012)Bruni, Boleda, Baroni, and
  Tran}]{bruni_distributional_2012}
Elia Bruni, Gemma Boleda, Marco Baroni, and Nam-Khanh Tran. 2012.
\newblock \href {https://aclanthology.org/P12-1015} {Distributional semantics
  in technicolor}.
\newblock In \emph{Proceedings of the 50th Annual Meeting of the Association
  for Computational Linguistics (Volume 1: Long Papers)}, pages 136--145.

\bibitem[{Caliskan et~al.(2017)Caliskan, Bryson, and Narayanan}]{caliskan2017}
Aylin Caliskan, Joanna~J Bryson, and Arvind Narayanan. 2017.
\newblock \href {https://doi.org/10.1126/science.aal4230} {Semantics derived
  automatically from language corpora contain human-like biases}.
\newblock \emph{Science}, 356(6334):183--186.

\bibitem[{Card et~al.(2020)Card, Henderson, Khandelwal, Jia, Mahowald, and
  Jurafsky}]{card2020}
Dallas Card, Peter Henderson, Urvashi Khandelwal, Robin Jia, Kyle Mahowald, and
  Dan Jurafsky. 2020.
\newblock \href {https://doi.org/10.18653/v1/2020.emnlp-main.745} {With little
  power comes great responsibility}.
\newblock In \emph{Proceedings of the 2020 Conference on Empirical Methods in
  Natural Language Processing (EMNLP)}, pages 9263--9274.

\bibitem[{Charlesworth et~al.(2022)Charlesworth, Caliskan, and
  Banaji}]{charlesworth2022}
Tessa~ES Charlesworth, Aylin Caliskan, and Mahzarin~R Banaji. 2022.
\newblock \href {https://doi.org/10.1073/pnas.2121798119} {Historical
  representations of social groups across 200 years of word embeddings from
  google books}.
\newblock \emph{Proceedings of the National Academy of Sciences},
  119(28):e2121798119.

\bibitem[{Charlesworth and Hatzenbuehler(2024)}]{charlesworth2024}
Tessa~ES Charlesworth and Mark~L Hatzenbuehler. 2024.
\newblock \href {https://doi.org/10.1038/s41598-024-61044-z} {Mechanisms
  upholding the persistence of stigma across 100 years of historical text}.
\newblock \emph{Scientific Reports}, 14(1):11069.

\bibitem[{Choi(2024)}]{choi2024}
Jonathan~H Choi. 2024.
\newblock \href
  {https://heinonline.org/HOL/LandingPage?handle=hein.journals/uclr91&div=4&id=&page=}
  {Measuring clarity in legal text}.
\newblock \emph{University of Chicago Law Review}, 91:1.

\bibitem[{Davani et~al.(2023)Davani, Atari, Kennedy, and Dehghani}]{davani2023}
Aida~Mostafazadeh Davani, Mohammad Atari, Brendan Kennedy, and Morteza
  Dehghani. 2023.
\newblock \href {https://doi.org/10.1162/tacl_a_00550} {Hate speech classifiers
  learn normative social stereotypes}.
\newblock \emph{Transactions of the Association for Computational Linguistics},
  11:300--319.

\bibitem[{Davies(2012)}]{davies2012}
Mark Davies. 2012.
\newblock \href {https://www.euppublishing.com/doi/abs/10.3366/cor.2012.0024}
  {Expanding horizons in historical linguistics with the 400-million word
  corpus of historical american english}.
\newblock \emph{Corpora}, 7(2):121--157.

\bibitem[{Dror et~al.(2020)Dror, Peled-Cohen, Shlomov, and Reichart}]{dror2020}
Rotem Dror, Lotem Peled-Cohen, Segev Shlomov, and Roi Reichart. 2020.
\newblock \href {https://link.springer.com/book/10.1007/978-3-031-02174-9}
  {\emph{Statistical Significance Testing for Natural Language Processing}}.
\newblock Number~45 in Synthesis Lectures on Human Language Technologies.
  Springer Nature.

\bibitem[{Ethayarajh et~al.(2019{\natexlab{a}})Ethayarajh, Duvenaud, and
  Hirst}]{ethayarajh2018}
Kawin Ethayarajh, David Duvenaud, and Graeme Hirst. 2019{\natexlab{a}}.
\newblock \href {https://doi.org/10.18653/v1/P19-1315} {Towards understanding
  linear word analogies}.
\newblock In \emph{Proceedings of the 57th Annual Meeting of the Association
  for Computational Linguistics}, pages 3253--3262, Florence, Italy.
  Association for Computational Linguistics.

\bibitem[{Ethayarajh et~al.(2019{\natexlab{b}})Ethayarajh, Duvenaud, and
  Hirst}]{ethayarajh2019}
Kawin Ethayarajh, David Duvenaud, and Graeme Hirst. 2019{\natexlab{b}}.
\newblock \href {https://doi.org/10.18653/v1/P19-1166} {Understanding
  undesirable word embedding associations}.
\newblock In \emph{Proceedings of the 57th Annual Meeting of the Association
  for Computational Linguistics}, pages 1696--1705, Florence, Italy.
  Association for Computational Linguistics.

\bibitem[{Garg et~al.(2018)Garg, Schiebinger, Jurafsky, and Zou}]{garg2018}
Nikhil Garg, Londa Schiebinger, Dan Jurafsky, and James Zou. 2018.
\newblock \href {https://doi.org/10.1073/pnas.1720347115} {Word embeddings
  quantify 100 years of gender and ethnic stereotypes}.
\newblock \emph{Proceedings of the National Academy of Sciences},
  115(16):E3635--E3644.

\bibitem[{Golub and Van~Loan(2013)}]{golub2013}
Gene~H Golub and Charles~F Van~Loan. 2013.
\newblock \href
  {https://www.press.jhu.edu/books/title/10678/matrix-computations}
  {\emph{Matrix Computations}}.
\newblock JHU press.

\bibitem[{Hamilton et~al.(2016)Hamilton, Leskovec, and Jurafsky}]{hamilton2016}
William~L. Hamilton, Jure Leskovec, and Dan Jurafsky. 2016.
\newblock \href {https://doi.org/10.18653/v1/P16-1141} {Diachronic word
  embeddings reveal statistical laws of semantic change}.
\newblock In \emph{Proceedings of the 54th Annual Meeting of the Association
  for Computational Linguistics (Volume 1: Long Papers)}, pages 1489--1501,
  Berlin, Germany. Association for Computational Linguistics.

\bibitem[{Han et~al.(2018)Han, Gill, Spirling, and Cho}]{han_conditional_2018}
Rujun Han, Michael Gill, Arthur Spirling, and Kyunghyun Cho. 2018.
\newblock \href {https://doi.org/10.18653/v1/D18-1527} {Conditional word
  embedding and hypothesis testing via bayes-by-backprop}.
\newblock In \emph{Proceedings of the 2018 Conference on Empirical Methods in
  Natural Language Processing}, pages 4890--4895. Association for Computational
  Linguistics.

\bibitem[{Jurgens et~al.(2012)Jurgens, Mohammad, Turney, and
  Holyoak}]{jurgens_semeval-2012_2012}
David Jurgens, Saif Mohammad, Peter Turney, and Keith Holyoak. 2012.
\newblock \href {https://aclanthology.org/S12-1047} {{SemEval}-2012 task 2:
  Measuring degrees of relational similarity}.
\newblock In \emph{*{SEM} 2012: The First Joint Conference on Lexical and
  Computational Semantics – Volume 1: Proceedings of the main conference and
  the shared task, and Volume 2: Proceedings of the Sixth International
  Workshop on Semantic Evaluation ({SemEval} 2012)}, pages 356--364.
  Association for Computational Linguistics.

\bibitem[{Kim et~al.(2016)Kim, Hullman, Burgess, and Adar}]{kim2016}
Yea-Seul Kim, Jessica Hullman, Matthew Burgess, and Eytan Adar. 2016.
\newblock \href {https://aclanthology.org/D16-1114.pdf} {Simplescience: Lexical
  simplification of scientific terminology}.
\newblock In \emph{Proceedings of the 2016 Conference on Empirical Methods in
  Natural Language Processing}, pages 1066--1071.

\bibitem[{Knoche et~al.(2019)Knoche, Popović, Lemmerich, and
  Strohmaier}]{knoche_identifying_2019}
Markus Knoche, Radomir Popović, Florian Lemmerich, and Markus Strohmaier.
  2019.
\newblock \href {https://doi.org/10.1145/3342220.3343658} {Identifying biases
  in politically biased wikis through word embeddings}.
\newblock In \emph{Proceedings of the 30th {ACM} Conference on Hypertext and
  Social Media}, {HT} '19, pages 253--257. Association for Computing Machinery.
\newblock Event-place: Hof, Germany.

\bibitem[{Levy and Goldberg(2014{\natexlab{a}})}]{levy_linguistic_2014}
Omer Levy and Yoav Goldberg. 2014{\natexlab{a}}.
\newblock \href {https://doi.org/10.3115/v1/W14-1618} {Linguistic regularities
  in sparse and explicit word representations}.
\newblock In \emph{Proceedings of the Eighteenth Conference on Computational
  Natural Language Learning}, pages 171--180. Association for Computational
  Linguistics.

\bibitem[{Levy and Goldberg(2014{\natexlab{b}})}]{levy2014}
Omer Levy and Yoav Goldberg. 2014{\natexlab{b}}.
\newblock \href
  {https://proceedings.neurips.cc/paper/2014/hash/feab05aa91085b7a8012516bc3533958-Abstract.html}
  {Neural word embedding as implicit matrix factorization}.
\newblock \emph{Advances in neural information processing systems}, 27.

\bibitem[{Levy et~al.(2015)Levy, Goldberg, and Dagan}]{levy2015}
Omer Levy, Yoav Goldberg, and Ido Dagan. 2015.
\newblock \href {https://doi.org/10.1162/tacl_a_00134} {Improving
  distributional similarity with lessons learned from word embeddings}.
\newblock \emph{Transactions of the Association for Computational Linguistics},
  3:211--225.

\bibitem[{Liao et~al.(2021)Liao, Taori, Raji, and Schmidt}]{liao2021we}
Thomas Liao, Rohan Taori, Inioluwa~Deborah Raji, and Ludwig Schmidt. 2021.
\newblock \href
  {https://datasets-benchmarks-proceedings.neurips.cc/paper/2021/hash/757b505cfd34c64c85ca5b5690ee5293-Abstract-round2.html}
  {Are we learning yet? {A} meta review of evaluation failures across machine
  learning}.
\newblock In \emph{Thirty-fifth Conference on Neural Information Processing
  Systems Datasets and Benchmarks Track (Round 2)}.

\bibitem[{Linzen(2016)}]{linzen_issues_2016}
Tal Linzen. 2016.
\newblock \href {https://doi.org/10.18653/v1/W16-2503} {Issues in evaluating
  semantic spaces using word analogies}.
\newblock In \emph{Proceedings of the 1st Workshop on Evaluating Vector-Space
  Representations for {NLP}}, pages 13--18. Association for Computational
  Linguistics.

\bibitem[{Liu et~al.(2023)Liu, Wu, Zhai, Yuan, and Zhang}]{liu2023}
Han Liu, Yuhao Wu, Shixuan Zhai, Bo~Yuan, and Ning Zhang. 2023.
\newblock \href
  {https://openaccess.thecvf.com/content/CVPR2023/html/Liu_RIATIG_Reliable_and_Imperceptible_Adversarial_Text-to-Image_Generation_With_Natural_Prompts_CVPR_2023_paper.html?ref=https://githubhelp.com}
  {Riatig: Reliable and imperceptible adversarial text-to-image generation with
  natural prompts}.
\newblock In \emph{Proceedings of the IEEE/CVF Conference on Computer Vision
  and Pattern Recognition}, pages 20585--20594.

\bibitem[{Lucy et~al.(2020)Lucy, Demszky, Bromley, and Jurafsky}]{lucy2020}
Li~Lucy, Dorottya Demszky, Patricia Bromley, and Dan Jurafsky. 2020.
\newblock \href {https://doi.org/10.1177/2332858420940312} {Content analysis of
  textbooks via natural language processing: Findings on gender, race, and
  ethnicity in texas us history textbooks}.
\newblock \emph{AERA Open}, 6(3):2332858420940312.

\bibitem[{Markovsky(2012)}]{markovsky2012}
Ivan Markovsky. 2012.
\newblock \href {https://link.springer.com/book/10.1007/978-1-4471-2227-2}
  {\emph{Low rank approximation: {A}lgorithms, implementation, applications}},
  volume 906 of \emph{Communications and Control Engineering}.
\newblock Springer.

\bibitem[{Matthews et~al.(2022)Matthews, Hudzina, and Sepehr}]{matthews2022}
Sean Matthews, John Hudzina, and Dawn Sepehr. 2022.
\newblock \href {https://doi.org/10.1609/aaai.v36i11.21461} {Gender and racial
  stereotype detection in legal opinion word embeddings}.
\newblock In \emph{Proceedings of the AAAI Conference on Artificial
  Intelligence}, volume~36, pages 12026--12033.

\bibitem[{Mikolov et~al.(2013{\natexlab{a}})Mikolov, Chen, Corrado, and
  Dean}]{mikolov_efficient_2013}
Tomas Mikolov, Kai Chen, Greg Corrado, and Jeffrey Dean. 2013{\natexlab{a}}.
\newblock \href {https://doi.org/10.48550/ARXIV.1301.3781} {Efficient
  estimation of word representations in vector space}.
\newblock Arxiv:1301.3781, version Number: 3.

\bibitem[{Mikolov et~al.(2013{\natexlab{b}})Mikolov, Yih, and
  Zweig}]{mikolov2013}
Tom{\'a}{\v{s}} Mikolov, Wen-tau Yih, and Geoffrey Zweig. 2013{\natexlab{b}}.
\newblock \href {https://aclanthology.org/N13-1090.pdf} {Linguistic
  regularities in continuous space word representations}.
\newblock In \emph{Proceedings of the 2013 Conference of the North American
  Chapter of the Association for Computational Linguistics: Human Language
  Technologies}, pages 746--751.

\bibitem[{Mimno and Thompson(2017)}]{mimno2017}
David Mimno and Laure Thompson. 2017.
\newblock \href {https://doi.org/10.18653/v1/D17-1308} {The strange geometry of
  skip-gram with negative sampling}.
\newblock In \emph{Proceedings of the 2017 Conference on Empirical Methods in
  Natural Language Processing}, pages 2873--2878, Copenhagen, Denmark.
  Association for Computational Linguistics.

\bibitem[{{National Archives at San
  Francisco}()}]{national_archives_case_nodate}
{National Archives at San Francisco}.
\newblock \href
  {https://drive.google.com/file/d/1ChwQCfbGr2P-P0JZNOGQhfohy20koq8f/view}
  {Case files by birth country}.

\bibitem[{Newberry et~al.(2017)Newberry, Ahern, Clark, and
  Plotkin}]{newberry2017}
Mitchell~G Newberry, Christopher~A Ahern, Robin Clark, and Joshua~B Plotkin.
  2017.
\newblock \href {https://doi.org/10.1038/nature24455} {Detecting evolutionary
  forces in language change}.
\newblock \emph{Nature}, 551(7679):223--226.

\bibitem[{Ng et~al.(2015)Ng, Allore, Trentalange, Monin, and Levy}]{ng2015}
Reuben Ng, Heather~G Allore, Mark Trentalange, Joan~K Monin, and Becca~R Levy.
  2015.
\newblock \href {https://doi.org/10.1371/journal.pone.0117086} {Increasing
  negativity of age stereotypes across 200 years: Evidence from a database of
  400 million words}.
\newblock \emph{PloS one}, 10(2):e0117086.

\bibitem[{Pennington et~al.(2014)Pennington, Socher, and
  Manning}]{pennington2014}
Jeffrey Pennington, Richard Socher, and Christopher~D Manning. 2014.
\newblock \href {https://aclanthology.org/D14-1162.pdf} {{GloVe}: Global
  vectors for word representation}.
\newblock In \emph{Proceedings of the 2014 conference on empirical methods in
  natural language processing (EMNLP)}, pages 1532--1543.

\bibitem[{Pennington et~al.(2023)Pennington, Socher, and
  Manning}]{pennington2014_gloverepo}
Jeffrey Pennington, Richard Socher, and Christopher~D Manning. 2023.
\newblock \href {https://github.com/stanfordnlp/GloVe} {{GloVe}: Global vectors
  for word representation}.
\newblock Github repository: Commit a6f2b94.

\bibitem[{Romano and Wolf(2017)}]{romano2017}
Joseph~P Romano and Michael Wolf. 2017.
\newblock \href {https://doi.org/10.1016/j.jeconom.2016.10.003} {Resurrecting
  weighted least squares}.
\newblock \emph{Journal of Econometrics}, 197(1):1--19.

\bibitem[{Sevim et~al.(2023)Sevim, {\c{S}}ahinu{\c{c}}, and
  Ko{\c{c}}}]{sevim2023}
Nurullah Sevim, Furkan {\c{S}}ahinu{\c{c}}, and Aykut Ko{\c{c}}. 2023.
\newblock \href {https://doi.org/10.1017/S1351324922000122} {Gender bias in
  legal corpora and debiasing it}.
\newblock \emph{Natural Language Engineering}, 29(2):449--482.

\bibitem[{Swinger et~al.(2019)Swinger, De-Arteaga, Heffernan~IV, Leiserson, and
  Kalai}]{swinger2019}
Nathaniel Swinger, Maria De-Arteaga, Neil~Thomas Heffernan~IV, Mark~DM
  Leiserson, and Adam~Tauman Kalai. 2019.
\newblock \href {https://doi.org/10.1145/3306618.3314270} {What are the biases
  in my word embedding?}
\newblock In \emph{Proceedings of the 2019 AAAI/ACM Conference on AI, Ethics,
  and Society}, AIES '19, page 305–311, New York, NY, USA. Association for
  Computing Machinery.

\bibitem[{Tang et~al.(2023)Tang, LeBel, Jain, and Huth}]{tang2023}
Jerry Tang, Amanda LeBel, Shailee Jain, and Alexander~G Huth. 2023.
\newblock \href {https://doi.org/10.1038/s41593-023-01304-9} {Semantic
  reconstruction of continuous language from non-invasive brain recordings}.
\newblock \emph{Nature Neuroscience}, 26(5):858--866.

\bibitem[{Tanner(1996)}]{tanner_tools_1996}
Martin~A. Tanner. 1996.
\newblock \href {https://doi.org/10.1007/978-1-4612-4024-2} {\emph{Tools for
  Statistical Inference: Methods for the Exploration of Posterior Distributions
  and Likelihood Functions}}, 3 edition.
\newblock Springer Series in Statistics. Springer New York.

\bibitem[{Ulmer et~al.(2022)Ulmer, Bassignana, M{\"u}ller-Eberstein, Varab,
  Zhang, van~der Goot, Hardmeier, and Plank}]{ulmer_experimental_2022}
Dennis Ulmer, Elisa Bassignana, Max M{\"u}ller-Eberstein, Daniel Varab, Mike
  Zhang, Rob van~der Goot, Christian Hardmeier, and Barbara Plank. 2022.
\newblock \href {https://doi.org/10.18653/v1/2022.findings-emnlp.196}
  {Experimental standards for deep learning in natural language processing
  research}.
\newblock In \emph{Findings of the Association for Computational Linguistics:
  EMNLP 2022}, pages 2673--2692, Abu Dhabi, United Arab Emirates. Association
  for Computational Linguistics.

\bibitem[{van~der Vaart(2000)}]{van2000}
Aad~W. van~der Vaart. 2000.
\newblock \href
  {https://www.cambridge.org/core/books/asymptotic-statistics/A3C7DAD3F7E66A1FA60E9C8FE132EE1D}
  {\emph{Asymptotic statistics}}, volume~3.
\newblock Cambridge University Press.

\bibitem[{Wendlandt et~al.(2018)Wendlandt, Kummerfeld, and
  Mihalcea}]{wendlandt2018}
Laura Wendlandt, Jonathan~K. Kummerfeld, and Rada Mihalcea. 2018.
\newblock \href {https://doi.org/10.18653/v1/N18-1190} {Factors influencing the
  surprising instability of word embeddings}.
\newblock In \emph{Proceedings of the 2018 Conference of the North {A}merican
  Chapter of the Association for Computational Linguistics: Human Language
  Technologies, Volume 1 (Long Papers)}, pages 2092--2102, New Orleans,
  Louisiana. Association for Computational Linguistics.

\bibitem[{Xiao et~al.(2023)Xiao, Baes, Vylomova, and Haslam}]{xiao2023}
Yu~Xiao, Naomi Baes, Ekaterina Vylomova, and Nick Haslam. 2023.
\newblock \href {https://doi.org/10.1371/journal.pone.0288027} {Have the
  concepts of ‘anxiety’ and ‘depression’ been normalized or
  pathologized? {A} corpus study of historical semantic change}.
\newblock \emph{PLOS one}, 18(6):e0288027.

\end{thebibliography}

\newpage 

\appendix

\section{Appendix}

\subsection{Training details}\label{app:training}
We trained multiple GloVe models for the experiments presented in Section \ref{sec:results}. First, we trained a single model on the unaltered COHA (1900-1999) corpus, which we used for all of the GloVe-V results. Second, we trained 100 GloVe models on document-level bootstrap samples of the COHA (1900-1999) corpus. In both cases, we pre-processed the corpus to lowercase all tokens and drop non-alphabetic characters.

We trained each 300-dimensional GloVe model (132M parameters for a vocabulary size of approximately 219,000 words) for 80 iterations using the following default hyperparameters of the official GloVe model implementation \citep{pennington2014_gloverepo}: initial learning rate of 0.05, and $\alpha=0.75$ and $x_{\max}=100$ for the weighting function. The training of each model consumed about 40 minutes on a workstation equipped with an AMD Milan 7543 @ 2.75 GHz processor CPU, using 48 CPU cores. 

\subsection{Variance Estimator Derivation for Bias Measures Using the Delta Method}
\label{app:delta}

\subsubsection{Cosine Similarity Bias}

The cosine similarity bias function of \citet{garg2018} is the following: 
\begin{align*}
\begin{split}
&f(\bs{v}_1, \dots, \bs{v}_k, \bs{M}_A, \bs{M}_W) = \\
    &\frac{1}{K}\sum_i \cos(\bs{v}_i, \bs{M}_A) - 
    \frac{1}{K} \sum_i  \cos(\bs{v}_i, \bs{M}_W)
\end{split}
\end{align*}
where $\bs{v}_i$ is the word vector for otherization word $i$, $\bs{M}_A$ is the mean word vector over all Asian surnames (using pre-normalized vectors), and $\bs{M}_W$ is the mean word vector over all White surnames (using pre-normalized vectors). There are $K$ otherization words.

In general, the partial derivative of the cosine similarity between two vectors $\bs{a}$ and $\bs{b}$ with respect to $\bs{a}$ is: 
\begin{align*}
    \frac{\partial \cos(\bs{a}, \bs{b})}{\partial \bs{a}} = \frac{\partial \bs{a}^T\bs{b}/\partial \bs{a}}{\|\bs{b}\| \|\bs{a}\|} - \frac{\cos(\bs{a}, \bs{b})}{\|
    \bs{a}\|} \cdot \partial \|\bs{a}\|/ \partial \bs{a}
\end{align*}
If we take $\bs{a}$ to be an otherization word $\bs{v}_i$, this gives us: 
\begin{align*}
\begin{split}
&\frac{\partial \cos(\bs{v}_i, \bs{M}_A)}{\partial \bs{v}_i} = \\ &\frac{\bs{M}_A}{\|\bs{M}_A\|\|\bs{v}_i\|} - \cos(\bs{v}_i, \bs{M}_A) \cdot \frac{\bs{v}_i}{\|\bs{v}_i\|^2}
\end{split}
\end{align*}

If we take $\bs{a}$ to be an Asian surname vector $\bs{a}_j$, where $\bs{M}_A = \frac{1}{m}\sum_j$ $\frac{\bs{a}_j}{||\bs{a}_j||}$, this gives us: 
\begin{align*}
\begin{split}
 &\frac{\partial \cos(\bs{v}_i, \bs{M}_A)}{\partial \bs{a}_j} =  \frac{1}{m}\bigg(\frac{I}{||\bs{a}_j||} - 
\frac{\bs{a}_j \bs{a}_j^T}{||\bs{a}_j||^3} \bigg) \\ 
&\left[\frac{\bs{v}_i}{||\bs{v}_i||||\bs{M}_A||} - \cos(\bs{v}_i, \bs{M}_A) \cdot \frac{\bs{M}_A}{||\bs{M}_A||^2} \right] \\
&= \frac{1}{m ||\bs{a}_j||||\bs{M}_A||}\bs{X}_{a_j}^T(\bs{\tilde{v}_i} - \cos(\bs{v}_i, \bs{M}_A) \cdot \bs{\tilde{M}_A})
\end{split}
\end{align*}
where $\bs{X}_{a_j} = \bs{I} - \bs{\tilde{a}}_j\bs{\tilde{a}}_j^T$.

So the partial derivatives with respect to each type of vector are the following:
\begin{align*}
\begin{split}
    &\frac{\partial f}{\partial \bs{v}_i} = \frac{1}{K||\bs{v}_i||} \\
    &\left( [\bs{\tilde{M}}_A -  \bs{\tilde{M}}_W] - \left[ \cos(\bs{v}_i, \bs{M}_A)  - 
    \cos(\bs{v}_i, \bs{M}_W) \right] \cdot \bs{\tilde{v}}_i \right) \coloneqq \bs{d}_v  \\
    &\frac{\partial f}{\partial \bs{a}_j} = 
    \frac{1}{Km||\bs{M}_A||} 
    \bigg(\frac{I}{||\bs{a}_j||} - \frac{\bs{a}_j \bs{a}_j^T}{||\bs{a}_j||^3} \bigg) \\
    &\sum_i \left( \bs{\tilde{v}}_i - \cos(\bs{v}_i, \bs{M}_A) \cdot \bs{\tilde{M}}_A \right) \\
    &= \frac{1}{Km ||\bs{a}_j||||\bs{M}_A||}\bs{X}_{a_j}^T \\
    &\sum_i\left( \bs{\tilde{v}}_i - \cos(\bs{v}_i, \bs{M}_A) \cdot \bs{\tilde{M}}_A \right) \coloneqq \bs{d}_a \\
   &\frac{\partial f}{\partial \bs{w}_j} = 
   -\frac{1}{Kn||\bs{M}_W||} 
   \bigg(\frac{I}{||\bs{w}_j||} - \frac{\bs{w}_j \bs{w}_j^T}{||\bs{w}_j||^3} \bigg) \\
   &\sum_i \left( \bs{\tilde{v}}_i - \cos(\bs{v}_i, \bs{M}_W) \cdot \bs{\tilde{M}}_W \right) \\
   &=  -\frac{1}{Kn ||\bs{w}_j||||\bs{M}_W||}\bs{X}_{w_j}^T \\
   &\sum_i\left( \bs{\tilde{v}}_i - \cos(\bs{v}_i, \bs{M}_W) \cdot \bs{\tilde{M}}_W \right) \coloneqq \bs{d}_w
\end{split}
\end{align*}

where $\bs{\tilde{a}}$ is the normalized version of vector $\bs{a}$.

\subsubsection{WEAT effect size}
\label{app:weat}

For two sets of attributes, $\bs{V}$ and $\bs{Z}$, of equal size ($k$) the WEAT effect size of \citet{caliskan2017} is the following: 

\begin{align*}
\begin{split}
&f(\bs{v}_1, \dots, \bs{v}_k, \bs{z}_1, \dots, \bs{z}_k, \bs{a}_1, \dots, \bs{a}_{A}, \bs{w}_1, \dots, \bs{w}_{W}) \\
&= \frac{\frac{1}{|V|} \sum_i s(\bs{v}_i, \bs{W}, \bs{A} ) -  \frac{1}{|Z|}\sum_i  s(\bs{z}_i, \bs{W}, \bs{A} )}{\text{std. dev}_{x \in V\cup Z} s(x, W, A)} \coloneqq \frac{H}{G}  
\end{split}
\end{align*}
where 
\begin{align*}
\begin{split}
&H = \sum_i \bigg( \frac{1}{W} \sum_j \cos(v_i, w_j) - \frac{1}{A} \sum_j \cos(v_i, a_j) \bigg) \\
&-  \sum_i \bigg( \frac{1}{W} \sum_j \cos(z_i, w_j) - \frac{1}{A} \sum_j \cos(z_i, a_j) \bigg) \nonumber \\
&= \sum_{v \in V} s(v, W, A) - \sum_{z \in Z} s(z, W, A)
\end{split}
\end{align*}
and
\begin{align*}
\begin{split}
    &G^2 =
    \frac{1}{|V \cup Z| - 1} \\
    &\sum_{x \in V \cup Z} \left( s(x, W, A) - \frac{1}{|Z \cup V|} \sum_{y} s(y, W, A) \right)^2 \\
    &\coloneqq \frac{1}{|V \cup Z| - 1} \sum_{x \in V \cup Z} \left( s(x, W, A) - E \right)^2
\end{split}
\end{align*}

Let $c_{\bs{a}}'(\bs{a}, \bs{b}) = \frac{\partial \cos(\bs{a}, \bs{b})}{\partial \bs{a}}$. We first define the following derivatives:

\begin{align*}
\begin{split}
     &\frac{\partial s(\bs{v}_i, W, A)}{\partial \bs{v}_i} = \frac{1}{W} \sum_j c_{\bs{v}_i}'(\bs{v}_i, \bs{w}_j) - \\ &\frac{1}{A} \sum_j c_{\bs{v}_i}'(\bs{v}_i, \bs{a}_j) \\
    &\frac{\partial s(\bs{z}_i, W, A)}{\partial \bs{z}_i} = \frac{1}{W} \sum_j c_{\bs{z}_i}'(\bs{z}_i, \bs{w}_j) - \\
    &\frac{1}{A} \sum_j c_{\bs{z}_i}'(\bs{z}_i, \bs{a}_j) \\
    &\frac{\partial s(\bs{x}, W, A)}{\partial \bs{w}_j} = 
   \frac{1}{W} c_{\bs{w}_j}'(\bs{w}_j, \bs{x}) \text{ , for } \bs{x} \in \{\bs{v}_i, \bs{z}_i\}\\
    &\frac{\partial s(\bs{x}, W, A)}{\partial \bs{a}_j} = 
    \frac{-1}{A} c_{\bs{a}_j}'(\bs{a}_j, \bs{x}) \text{ , for } \bs{x} \in \{\bs{v}_i, \bs{z}_i\} 
\end{split}
\end{align*}

The partial derivatives with respect to each type of vector are the following: 
\begin{align*}
\begin{split}
&\frac{\partial f}{\partial \bs{v}_i} =  \\
& \left(\frac{1}{G|V|} - \frac{H}{G^{\frac{5}{2}}} \cdot \frac{\left( s(\bs{v}_i, W, A) - E \right)}{|V \cup Z|} \right) \frac{\partial s(\bs{v}_i, W, A)}{\partial \bs{v}_i} \\
&\frac{\partial f}{\partial \bs{z}_i} =  \\
&\left(\frac{-1}{G|Z|}  - \frac{H}{G^{\frac{5}{2}}} \cdot \frac{ \left( s(\bs{z}_i, W, A) - E \right)}{|V \cup Z|} \right) \frac{\partial s(\bs{z}_i, W, A)}{\partial \bs{z}_i} \\
&\frac{\partial f}{\partial \bs{w}_j} = 
    \frac{1}{G} \cdot \\
    &\left(\frac{1}{|V|} \sum_v \frac{\partial s(\bs{v}_i, W, A)}{\partial \bs{w}_j} - \frac{1}{|Z|} \sum_z \frac{\partial s(\bs{z}_i, W, A)}{\partial \bs{w}_j} \right) \\
&- 
    \frac{H}{G^{\frac{5}{2}}} \frac{1}{|V \cup Z| - 1} \\
    &\resizebox{0.5\textwidth}{!}{$\sum_{x \in V \cup Z} \left( ( s(\bs{x}, W, A) - E ) \cdot \left(\frac{\partial s(\bs{x}, W, A)}{\partial \bs{w}_j} - \frac{\partial E}{\partial \bs{w}_j} \right) \right)$}
     \\
   &\frac{\partial f}{\partial \bs{a}_j} = 
   \frac{1}{G} \cdot \\
   &\left(\frac{1}{|V|} \sum_v \frac{\partial s(\bs{v}_i, W, A)}{\partial \bs{a}_j} - \frac{1}{|Z|} \sum_z \frac{\partial s(\bs{z}_i, W, A)}{\partial \bs{a}_j} \right) \\
    &- 
    \frac{H}{G^{\frac{5}{2}}} \frac{1}{|V \cup Z| - 1} \\
    &\resizebox{0.5\textwidth}{!}{$\sum_{x \in V \cup Z} \left( ( s(\bs{x}, W, A) - E ) \cdot \left(\frac{\partial s(\bs{x}, W, A)}{\partial \bs{a}_j} - \frac{\partial E}{\partial \bs{a}_j} \right) \right)$}
\end{split}
\end{align*}
where 
\begin{align*}
    \frac{\partial E}{\partial \bs{w}_j} &= \frac{1}{|V \cup Z|} \sum_{x \in V \cup Z} \frac{\partial s(\bs{x}, W, A)}{\partial \bs{w}_j} \\
    \frac{\partial E}{\partial \bs{a}_j} &= \frac{1}{|V \cup Z|} \sum_{x \in V \cup Z} \frac{\partial s(\bs{x}, W, A)}{\partial \bs{a}_j}
\end{align*}

\subsubsection{Delta Method}

Using the derivatives computed in the sections above, the variance of the bias calculation is:
\begin{align*}
    \text{var}(h) = \sum_i (\bs{d}_t)_i^T \bs{\Sigma}_i(\bs{d}_t)_i
\end{align*}
where $h \in [f, g]$ is the bias function,
$t$ is the type of word $i$ (e.g., $t = a$ for Asian surnames, $t= w$ for White surnames, $t=v$ for Otherization words in the case of the cosine similarity metric), and $\bs{\Sigma}_i$ is the variance-covariance matrix for the parameters of word $i$ (Equation \ref{eq:sigma}).

\subsection{Asian Surname List Generation}
\label{app:surname}
To explore the behavior of anti-Asian bias scores on words with varying frequencies in the COHA (1900--1999) corpus (Section \ref{sec:bias}), we compile a novel Asian surname list with the objective of capturing a broader and more representative set of surnames that would be present in a historic corpus such as COHA. 

We build on two existing and widely used surname lists for ethnic bias measurement. First, a list of 20 Asian last names curated by \citet{garg2018}. This was designed to include the most common surnames in the United States for this ethnicity, as measured by 2000 Census data, as well as the surnames that had higher average frequencies in the Google Books and COHA corpora studied by the authors (largely covering the 1800--1999 period). As a result of this curation process, the list is solely focused on higher frequency last names, and, by primarily comprising Chinese surnames, may not be wholly representative of the Asian ethnicity and of the historic appearances of Asian surnames in this corpus. Second, a list of 200 Asian Pacific Islander surnames collected by \citet{matthews2022} from the 2010 U.S. decennial census, sampled from last names that had a probability of 90\% or larger of belonging to this ethnicity. 

We expand the set of 211 unique Asian last names from the \citet{garg2018} and \citet{matthews2022} lists by collecting surnames in immigration arrival cases from the National Archives at San Francisco, California, from 1910--1940 \citep{national_archives_case_nodate}. This data includes over 65,000 cases detailing the country of birth, arrival date, first and last name, gender and date of birth of each person. As ethnicity is not directly reported, we use the country of birth to compile the Asian surname list, and find over 1,300 unique last names belonging to immigrants whose reported birthplace is one the following locations: China, Japan, Korea, Indo-China, India, Hong Kong, Burma, Philippine Islands, Thailand, Malaysia, and Mongolia.\footnote{We make this surname list available to researchers in our code repository.}

\subsection{Alternative Approaches to Word Embedding Estimation Uncertainty}
\label{app:alt}

We explored several approaches to measuring the reconstruction error of word embeddings, in addition to the probabilistic model for GloVe that we presented in this work. 

\subsubsection{Implicit matrix factorization in the skip-gram with negative-sampling (SGNS) model}
Per \citet{levy2014}, the SGNS word embedding model implicitly factorizes a matrix that contains the shifted pointwise mutual information (PMI) of word and context vectors. Let $\bs{w}$ be a word vector for word $w$, $\bs{c}$ be a context vector for word $c$, $V_W$ and $V_C$ be the word and context vocabularies, respectively, $k$ be the number of negative samples, $D$ be the collection of word and context pairs in a corpus, $\#(i)$ be the number of occurrences of word $i$ in the corpus, and $\#(i, j)$ the number of co-occurrences of the words $i$ and $j$ in the corpus. Then, for sufficiently large dimensionality of the embeddings, the optimal vectors according to the SGNS objective are such that:
\begin{align*}
    \bs{w} \cdot \bs{c} &= \log \left(\frac{\#(w, c) \cdot |D|}{\#(w) \cdot \#(c)}\right) - \log k \\
    &= PMI(w, c) - \log k
\end{align*}

In this manner, a measure of error in the estimation of the word embedding for word $w$ could be obtained by comparing the dot product $\bs{w} \cdot \bs{c}$ and $PMI(w, c)$ across all context words $c$ appearing with word $w$ in the corpus. In particular, we explored a word-level measure of estimation error for word $w$ captured by the median of $d(w, c)$, the context-level percentage of deviation from the optimal value, over all contexts: 
\begin{align*}
    d(w, c) &=  \frac{\bs{w} \cdot \bs{c} - (PMI (w, c) - \log k)}{PMI(w, c) - \log k}
\end{align*}

\begin{figure}[t]
    \centering
    \includegraphics[ width=\columnwidth]{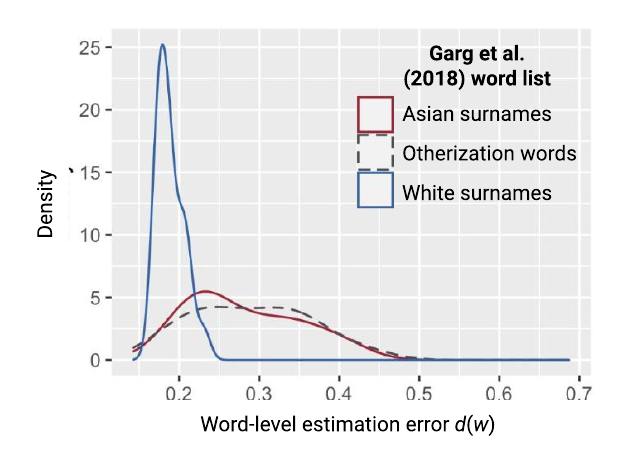}
    \caption{\textbf{Distribution of word-level estimation error measures in COHA for different word lists.} Using \citeauthor{levy2014}'s findings for the skip-gram with negative-sampling (SGNS) model, we compute a word-level measure of the estimation error in word embeddings.  }
    \label{fig:app_SPMI}
\end{figure}

By comparing the distribution of this word-level measure of estimation error over different word lists, we found that higher frequency word lists, such as the set of White surnames of \citet{garg2018}, had much lower estimation errors compared to relatively lower frequency word lists, such as the set of 20 Asian surnames (see \figurename~\ref{fig:app_SPMI}). 

Though helpful to assess the estimation quality of word embeddings in a corpus, an important limitation of this method is that it is not a probabilistic model, and so offers no obvious way to perform hypothesis testing in downstream applications accounting for this type of error. For this reason, we sought a different approach that could provide not only point estimates for the embeddings, but accompanying uncertainty measures with a probabilistic foundation. 

\subsubsection{Bayes by Backprop for the SGNS model}
In order to obtain distributions over the word embeddings, we explored the Bayes-by-Backprop algorithm of \citet{blundell_weight_2015}, a variational inference approach that learns a distribution over neural network weights. \citet{han_conditional_2018} adapt this method to obtain approximate posterior distributions for SGNS word embeddings, incorporating metadata on document covariates in order to learn conditional distributions for different corpus subsets (e.g., temporal periods or genres) that share structural information across these partitions. Using a Gaussian mixture prior for the parameters of the word and context vectors, this method computes the following conditional posterior distribution $\bs{w}_{w|x}$ for word vectors and unconditional posterior distribution $\bs{w}_c$ for the context vectors:
\begin{align*}
    \bs{w}_{w|x} &\sim N(f(\bs{\mu}_{w}, \bs{\mu}_x), \bs{\sigma}_{w|c}) \\
    \bs{w}_c &\sim N(\bs{\tilde{\mu}}_c, \tilde{\bs{\sigma}}_c))
\end{align*}
where $x$ is the subcorpus on which the embedding for word $w$ is estimated, $f$ is an affine transformation that combines corpus-level word vectors $\bs{\mu}_{w}$ and embeddings for each subcorpora $\bs{\mu}_x$, and $\bs{\sigma}_w$ and $\tilde{\bs{\sigma}}_c$ are the diagonal covariance of word and context vectors, respectively, parameterized as $\bs{\sigma}_{w|c} = \log(1 + e^{\bs{\rho}_w})$ and $\tilde{\bs{\sigma}}_c = \log(1 + e^{\bs{\tilde{\rho}}_c})$. The Bayes-by-Backprop algorithm initializes parameters $\bs{\mu}_{w}$, $\bs{\mu}_x$, $\bs{\tilde{\mu}}_c$, $\bs{\rho}_w$ and $\bs{\tilde{\rho}}_c$ for all word and context vectors in the vocabulary, and, given $(w, c, x)$ triplets, performs sequential updates to these parameters by computing the gradient of the variational approximation to the posterior.

\begin{figure}[t]
    \centering
    \includegraphics[ width=\columnwidth]{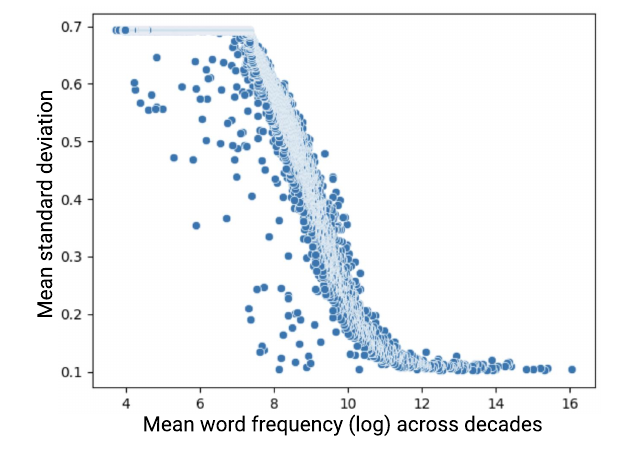}
    \caption{\textbf{Word-level relationship between posterior standard deviations and frequency in COHA using the Bayes-by-Backprop approach.} Lower-frequency words displayed poor convergence and their posterior standard deviations were effectively unchanged during training.   }
    \label{fig:app_bbb}
\end{figure}

Unfortunately, training meaningful embeddings with Bayes-by-Backprop turned out to be challenging. On the COHA corpus, our best performance was a mean accuracy of 0.11 on the Google analogy task \citep{mikolov_efficient_2013} and a mean Pearson similarity statistic of 0.41 on the MEN similarity task \citep{bruni_distributional_2012}, relative to a benchmark of 0.24 and 0.51, respectively, using the pre-trained embeddings of \citet{hamilton2016} across COHA decades. We explored numerous refinements to improve the training of Bayesian Neural Networks \citep{book_training_2020}, including using different weight initialization schemes for separate parameter groups (e.g., Uniform Kaiming scheme), and both uniformly and dynamically re-weighting the Kullback-Leibler divergence component of the cost function, to no greater success.

In addition to the lower quality of the embedding posterior means, we encountered an important scaling issue in the trained parameters. After training, the posterior means were one to two orders of magnitude smaller than the posterior standard deviations, effectively barring us from drawing meaningful samples from these distributions. An analysis of the relationship between the posterior standard deviations and word-level frequency indicated that there was an inverse relationship between these (similar to what we document in \figurename~\ref{fig:validationl2} for GloVe-V); however, this relationship only held for higher frequency words. For lower frequency words, the posterior standard deviations were all equivalent and effectively unmodified from their initialization value (see \figurename~\ref{fig:app_bbb}). Convergence diagnostics on this model confirmed that these parameters ($\bs{\rho}_w$ and $\bs{\tilde{\rho}}_c$) were not training correctly. Further work would be required to design priors and parameter-specific weight initialization schemes that lead to proper training of the parameters for this subset of words. Our code for these training runs is made available at the project repository.

\end{document}